# PatentTransformer-2
# Controlling Patent Text Generation by Structural Metadata

Jieh-Sheng Lee* and Jieh Hsiang

Department of Computer Science and Information Engineering
National Taiwan University, Taiwan
{d04922013, hsiang}@csie.ntu.edu.tw

## Abstract

PatentTransformer is our codename for patent text generation based on Transformer-based models. Our goal is "Augmented Inventing." In this second version, we leverage more of the structural metadata in patents. The structural metadata includes patent title, abstract, and dependent claim, in addition to independent claim previously. Metadata controls what kind of patent text for the model to generate. Also, we leverage the relation between metadata to build a text-to-text generation flow, for example, from a few words to a title, the title to an abstract, the abstract to an independent claim, and the independent claim to multiple dependent claims. The text flow can go backward because the relation is trained bidirectionally. We release our GPT-2 models trained from scratch and our code for inference so that readers can verify and generate patent text on their own. As for generation quality, we measure it by both ROUGE and Google Universal Sentence Encoder.

## 1 Introduction

**Auto-complete.** Our ultimate goal is to build an "Augmented Inventing" system to help inventors conceive better inventions. An envisioned use case is an "auto-complete" function in which, if an inventor is contemplating and has no whole picture in mind yet, patent claim generation can augment the inventor to explore relevant ideas from patents in the past. In our work (Lee and Hsiang, 2019b), we trained a GPT-2 model (Radrof et al., 2018) for generating independent claims. In our work (Lee and Hsiang, 2019c), we trained a BERT model (Devlin et al., 2019) for patent classification. In our work (Lee and Hsiang, 2019a), we reuse the skills to train a BERT model for measuring claim span relevancy. The purpose of span relevancy is to measure the text generation quality of a GPT-2 model. In my previous work (Lee, 2019), I analyzed the "auto-complete" function in four perspectives and proposed using inventor-centric data to make patent text generation more personalized. By then, PatentTransformer had become our codename to represent the first version of our workable system.

**Metadata and relation.** In this work, we expand the scope of text generation to cover patent title, abstract, and dependent claim, in addition to the previous independent claim. We treat these four types of patent text as structural metadata. We also identify the three most relevant structural metadata pairs: (title, abstract), (abstract, independent claim), (independent claim, dependent claim). The relation in a pair is bidirectional. We define text-to-text mapping based on the bidirectional relation. Such a bidirectional generation also applies to the text generation based on four structural metadata. Based on the above, our training data contains the patent text in four structural metadata and the text-to-text mapping in three metadata pairs. Collectively these seven kinds of data make GPT-2 text generation cover broader structural data with more fine-grained control.

---

* Admitted in New York and passed the USPTO patent bar exam. Currently a Ph.D. candidate focusing on Deep Learning for patents and an in-house patent counsel at Novatek Microelectronics Corp.

**Generation flow.** For example, a use case of text generation flow in PatentTransformer-2 is: (1) an inventor writes a few words as the seed text for a patent title, (2) the model generates the rest of the title forward and backward, (3) based on the title, the model generates a patent abstract, and (4) based on the abstract, the model generates its independent claim and dependent claim(s). Such a text generation flow is adjustable for different needs. One can start from different metadata, and bidirectionally, end at another desired metadata.

## 2 Related Work

For more related works in both the patent field and the computer science field, interested readers can refer to the same section in our previous works (Lee and Hsiang, 2019b; Lee and Hsiang, 2019c; Lee and Hsiang, 2019a; Lee, 2019). We update some recent progress by other researchers here. First, we cite (Keskar et al., 2019) to show the possibility of a conditional Transformer language model for controllable generation (CTRL). The model is trained to condition on control codes that govern style, content, and task-specific behavior. The model is always conditioned on the control codes to calculate the loss in the distribution of the language model. In this work, our idea of using control codes is similar, but the implementation is different. We treat the structural metadata as regular tokens like other text tokens and feed them together into a GPT-2 model. No special treatment on the control codes is required. Second, (Dathathri et al., 2019) proposed a Plug and Play language model as a simple approach to control text generation. The idea is to combine an attribute model to push the hidden activations of the language model with gradients through a forward and backward pass. How to integrate such an approach with our work is something we planned for the future.

## 3 Data

**More data and tags.** The raw data for training GPT-2 models from scratch is from the Google Patents Public Datasets on BigQuery.[1] We take the same and previous span-based approach. For details of the preprocessing in the data pipeline, please refer to our works (Lee and Hsiang, 2019b; Lee and Hsiang, 2019a). The main differences here are: (1) we collected more utility patents (1976~2017-08) from the data source, (2) we added more special tags to represent the structural metadata and the text-to-text mapping between two metadata. In Table 1, the two special tags in row 1 were defined in our previous work for claims. In PatentTransformer-2, we add row 2 to 6 as new tags to specify the title, the abstract, and the backward direction.

| Metadata | prefix | appendix |
|---|---|---|
| claim | <\|startoftext\|> | <\|endoftext\|> |
| | <\|startofbackward\|> | <\|endofbackward\|> |
| title | <\|startoftitle\|> | <\|endoftitle\|> |
| | <\|backwardtitlestart\|> | <\|backwardtitleend\|> |
| abstract | <\|startofabstract\|> | <\|endofabstract\|> |
| | <\|backwardabstractstart\|> | <\|backwardabstractend\|> |

Table 1. Special tags for metadata.

**Tag for relation.** Besides, the previous version handled independent claims only. Since there is only one metadata then, no relation exists between metadata. In this version, there are multiple metadata. We see a chance to leverage the relation introduced by new metadata when preparing training data. In Table 2, we define five special tags for text-to-text mapping and annotating the relation between metadata 1 and metadata 2. By training the model with the mapping relations, it is feasible to specify what kind of metadata the model should generate. For example, by appending the tag `<|title2abstract|>` to an abstract, the model is expected to generate an abstract. Based on experiences, the most direct relations are between: title and abstract, abstract and independent claim, independent claim and dependent claim. Row 2 to 6 in Table 2 captures such relations with new special tags.

---
[1] https://console.cloud.google.com/bigquery?p=patents-public-data

| Metadata 1 | Text-to-text mapping | Metadata 2 |
|---|---|---|
| independent claim | <\|dep\|> | dependent claim |
| dependent claim | <\|dep\|> | dependent claim |
| title | <\|title2abstract\|> | abstract |
| abstract | <\|abstract2claim\|> | claim |
| claim | <\|claim2abstract\|> | abstract |
| abstract | <\|abstract2title\|> | title |

Table 2. Special tags for text-to-text mapping.

**Size.** During pre-processing, each dependent claim concatenates with its depended claim as a record. Multiple claim dependencies are skipped since they are less frequent. After pre-processing the raw patent text with special tags in both forward and backward directions, the dataset covering 1976~2016 contains 371,731,396 records. The dataset covering 2016 contains 20,756,172 records. The dataset covering 2017-01~08 contains 12,488,254 records. The majority of the records are independent claims and dependent claims since it is common for a patent to have multiple claims. We use the datasets 1976~2016 and 2016 for training and the datasets 2016 and 2017 for testing.

**Dataset for TF.** For efficiency without repeated pre-processing, the datasets are converted to the native TF format for TensorFlow by randomizing and tokenizing text records. Each file in TF format contains 4,096 lines. If a tokenized record is longer than the context window (1,024) of the model, we take a sliding window approach to cover the rest of the record as multiple lines in the TF file. If a tokenized record is shorter, we append the record by randomly picking other records. The purpose is to utilize the computing resource for the whole context window during training.

## 4   Method & Experimental Setup

In this section, we provide our repository on GitHub, some repositories we leveraged or considered, and the model sizes in our experiments.

### 4.1   GitHub

**TPU & GPU.** Our code for PatentTransformer is released [2] for researchers. The "v2" directory contains the sample code in this work, and the "v1" directory contains our code in previous works. During our development, there are several repositories specific to GPT-2 available on GitHub. After evaluation, we leverage Leahy's repository[3] for training and OpenAI's repository[4] for inference. OpenAI trains their models with TPU, but the code for training was not released. To our knowledge, Leahy's repository is the first public repository capable of training a GPT-2 model on TPU. Therefore, we leverage the code and the free TPU on Google Colab. As for inferencing, we use the original code released by OpenAI which works with GPU. The model trained from scratch by Leahy's code is compatible with OpenAI's code for inference.

**Alternatives.** Later on, we found that HuggingFace's Transformers 2.0 [5] might be a better option for future works. The repository covers several state-of-the-art Transformer-based models, such as BERT, GPT-2, RoBERTa (Liu et al., 2019), XLM (Lample and Conneau, 2019), DistilBert (Sanh et al., 2019), XLNet (Yang et al., 2019), CTRL, etc., and provides over 32+ pre-trained models in 100+ languages and interoperability between TensorFlow 2.0 and PyTorch. Another repository using TPU is Grover,[6] which aims to build a state-of-the-art defense against neural fake news. Based on Grover, Zhang builds a 1.5B pre-trained Chinese model in the GPT2-ML (GPT2 for Multiple Languages) repository. [7] These repositories are what we plan to train larger and multilingual models in the future.

### 4.2   Model Sizes & Hyperparameters

**Compatible.** In this work, we trained a small model and a medium model from scratch for benchmarking and for proof of concept. We followed the hyperparameters provided by OpenAI, such as vocabulary size (50,257), context size (1024), embedding size (1024 for medium & 768 for small), etc. We reused the same BytePair Encoding settings (vocab.bpe & encoder.json) by OpenAI too so that our pre-trained models are compatible with most of the GPT-2-related repositories for inferencing or fine-tuning with different corpora. For text generation, we use the default top_k (40) in OpenAI's code. The number of training steps is one million and the batch size is 8. The number of warmup steps is 10,000. We use Adam as the optimizer and set the learning rate as 1e-4. In Leahy's code, dropout is added to the model and we set it as 0.1.

---
[2] https://github.com/jiehsheng/PatentTransformer
[3] https://github.com/ConnorJL/GPT2
[4] https://github.com/openai/gpt-2
[5] https://github.com/huggingface/transformers
[6] https://github.com/rowanz/grover
[7] https://github.com/imcaspar/gpt2-ml

Figure 1. Source code for text generation flow.

## 5 Results

First, we provide the sample code for the text generation flow from a few words to multiple dependent claims. Second, we show the training loss of our four models. Third, we compare model performance and explain how they are calculated.

### 5.1 Demo & Code

**Functions.** In this section, we demonstrate how to generate from a few words in the patent title to an abstract, from the generated abstract to an independent claim, and from the generated independent claim to multiple dependent claims in the end. Our code runs on Colab and is convenient for researchers to enhance and test. For end users, it might be preferable to wrap the code with different user interfaces. Figure 1 shows two kinds of high-level functions: (1) `patent_text_gen()` for generating text based on input text, metadata and the number of records to generate, and (2) `text2text_mapping()` for mapping the text from one kind of metadata to another.

**Step by step.** In the above example, "temperature optimization" is the seed text for the code to generate a patent title bidirectionally. The `outputs` variable is a list for receiving the generated text. In this case, there is only one record because of `gen_count=1`. Next, the `outputs[0]` (the generated patent title) is passed to `text2text_mapping()` function. The function can generate a patent abstract based on `mapping='title2abstract'`. Then, the `outputs[0]` (the generated patent abstract) is passed to `text2text_mapping()` function. The function can generate an independent claim based on `mapping='abstract2claim'`. Last, the `outputs[0]` (the generated independent claim) is passed to `text2text_mapping()` function. The function can generate two dependent claims based on `mapping='dep'` and `gen_count=2`. Supplement A[8] lists 100 examples (no cherry-picking) of such text generation flow, and the following is the "run 4" result in the list (independent claim and dependent claims are omitted here for saving space). It is noted that the sequence of "(a)…; (b)…; (c)…; and (d)…." in the example is correct.

Title: *Control method and temperature optimization for temperature compensation in a thermoelectric system having a plurality of semiconductor chips*

Abstract: *A control method and temperature optimization for temperature compensation in a thermoelectric system having a plurality of semiconductor chips mounted on a circuit board and at*

---
[8] https://github.com/jiehsheng/PatentTransformer/blob/master/v2/(paper)%20Supplement_A.txt

*least two thermal sensors connected to different locations on the at least one semiconductor chip using heat dissipative members to conduct or dissipate heat. The method including: (a) generating a first set of electrical signals corresponding to the respective temperatures to a first semiconductor chip of the semiconductor chips; (b) generating a second set of electrical signals by applying one of at least two electrical signals among the first set of electrical signals and the other electrical signal among the first set of electrical signals; (c) estimating the temperature corresponding to the temperature sensor of the second semiconductor chip by estimating the temperature of the thermal sensor connected to the thermal sensor based on the first electrical signal and the second electrical signal; and (d) setting the temperature corresponding to the thermal sensor of the first semiconductor chip based on a result of the estimation.*

## 5.2 Training Loss

**Four models.** On Colab, only the small and the medium model sizes are feasible for training. The Large and the extra-large models encountered an OOM (Out-of-Memory) issue in our experiments. We train both the small and medium models from scratch for one million steps. The data periods for our training datasets are 1976~2016 and 2016. In total, we train four models: a small model for 1976~2016 (M1), a medium model for 1976~2016 (M2), a small model for 2016 (M3), and a medium model for 2016 (M4). Figure 2 shows the training loss after one million steps for each of them, and the model M2 is our best result. It is noted that a lower training loss is not necessarily the best for downstream tasks.

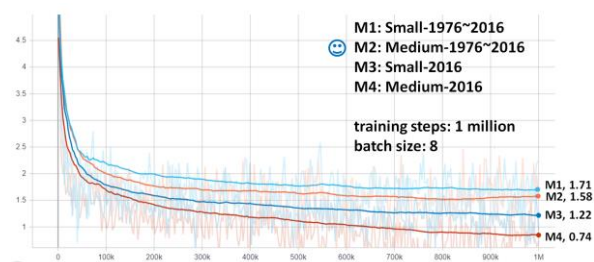

Figure 2: Training loss.

## 5.3 Model Performance

**Metrics.** We compare model performance by the conventional ROGUE metric and the semantic similarity measured by the Universal Sentence Encoder (version 2) released by Google.[9] It encodes text into high-dimensional vectors that can be used for text classification, semantic similarity, clustering, and other natural language tasks. During our manual inspection, we found the semantic similarity is a better metric for patent text. For example, the similarity of the following two titles is 95.32%, and the F1 score of ROUGE-1 is 63.16%:

*(predicted) Organic light emitting display unit structure*

*(actual) Organic light emitting display unit structure and organic light emitting display unit circuit*

**Result files.** In Table 3, we compare our four models (M1~M4) and their testing on the 2016 and 2017 datasets, concerning four text-to-text mappings: Abstract to Title, Title to Abstract, Abstract to Claim, Claim to Abstract. We test 1,000 records for each setting in the performance comparison and calculate the average of their ROUGE-1 values and Similarity values. For example, regarding Abstract to Title, we select 1,000 records in actual abstracts, feed each record for the model to generate a predicted title, then calculate the metric based on the predicted title and the actual title. In total, there are 32 settings in the table. We provide the details of testing for each setting in the Supplement C folder[10] in our GitHub repository. In Supplement B,[11] we pick 19 records showing the Similarity value ranging from 1.0, 0.95, 0.90, 0.85 till 0.11 as an easier way to judge the effectiveness of the metrics (setting: Abstract to Title, Model: Small, Train with 2016 data, Test with 2016 data).

## 6 Looking Forward

In the future version, we plan to work on: (1) how to generate patent text from scientific papers, (2) training the model to learn the legal requirements in patent law, such as novelty, utility, nonobviousness, written requirement, etc., and (3) inventor-centric patent generation. Any progress in directions like these would be a step closer to our ultimate goal: build an "Augmented

---

[9] https://tfhub.dev/google/universal-sentence-encoder/2

[10] https://github.com/jiehsheng/PatentTransformer/tree/master/v2/(paper)%20Supplement_C

[11] https://github.com/jiehsheng/PatentTransformer/blob/master/v2/(paper)%20Supplement_B.pdf

| Text-to-Text | Model Size | Train 1M steps | Test 2016 | | Test 2017 | |
|---|---|---|---|---|---|---|
| | | | ROUGE-1 (%) | Similarity (%) | ROUGE-1 (%) | Similarity (%) |
| Abstract to Title | Small | 2016 | F1: 40.32 P: 43.47 R: 45.20 | 70.54 | F1: 38.39 P: 42.25 R: 43.62 | 69.63 |
| | Small | 1976~2016 | F1: 39.33 P: 42.53 R: 45.01 | 70.85 | F1: 39.53 P: 43.03 R: 44.67 | 70.98 |
| | Medium | 2016 | F1: 38.75 P: 41.14 R: 45.49 | 70.08 | F1: 36.53 P: 39.82 R: 42.42 | 69.19 |
| | **Medium** | **1976~2016** | F1: 42.02 P: 44.87 R: 48.53 | **72.45** | F1: 40.02 P: 43.61 R: 45.19 | **70.91** |
| Title to Abstract | Small | 2016 | F1: 31.86 P: 32.48 R: 36.04 | 67.37 | F1: 31.76 P: 32.75 R: 35.53 | 67.57 |
| | Small | 1976~2016 | F1: 30.99 P: 31.12 R: 36.02 | 67.59 | F1: 31.70 P: 32.77 R: 35.77 | 67.90 |
| | Medium | 2016 | F1: 30.93 P: 32.85 R: 33.39 | 65.44 | F1: 30.27 P: 32.60 R: 32.79 | 65.86 |
| | **Medium** | **1976~2016** | F1: 32.22 P: 32.20 R: 37.25 | **68.69** | F1: 31.92 P: 31.71 R: 37.56 | **69.64** |
| Abstract to Claim | Small | 2016 | F1: 43.48 P: 46.96 R: 48.02 | 75.59 | F1: 42.63 P: 45.92 R: 47.56 | 74.73 |
| | Small | 1976~2016 | F1: 46.14 P: 51.34 R: 49.72 | 77.87 | F1: 46.00 P: 51.48 R: 49.75 | 78.22 |
| | Medium | 2016 | F1: 38.24 P: 42.48 R: 40.85 | 68.40 | F1: 34.99 P: 39.80 R: 36.77 | 66.30 |
| | **Medium** | **1976~2016** | F1: 47.24 P:53.37 R: 49.19 | **79.05** | F1: 48.32 P: 55.04 R: 50.06 | **79.72** |
| Claim to Abstract | Small | 2016 | F1: 46.47 P: 49.11 R: 49.45 | 79.82 | F1: 44.83 P: 47.27 R: 47.79 | 78.91 |
| | Small | 1976~2016 | F1: 45.03 P: 47.38 R: 49.14 | 79.54 | F1: 45.21 P: 48.27 R: 48.01 | 79.30 |
| | Medium | 2016 | F1: 42.25 P: 44.39 R: 45.30 | 76.47 | F1: 38.33 P: 40.84 R: 40.36 | 73.84 |
| | **Medium** | **1976~2016** | F1: 45.62 P: 47.64 R: 49.74 | **79.91** | F1: 45.93 P: 47.71 R: 49.75 | **80.27** |

Table 3. Model performance comparison.

Inventing" system to help humans be more creative.

## 7 Conclusion

In this work, we leverage the structural metadata in patents for controlling text generation and their relations for text-to-text mapping and generating a text flow. We also released four models trained from scratch and the sample code to demonstrate how to generate a patent title bidirectionally from a few words, an abstract from the title, an independent claim from the abstract, and multiple dependent claims from the independent claim. The performances of the models are measured by ROGUE and Universal Sentence Encoder for benchmarking. Conceptually, what works in this paper is not limited to patents and may work for other types of documents if similar structural metadata and metadata relation exist.

## Acknowledgments

We thank the anonymous reviewers for their valuable feedback. The research reported in this manuscript has been funded by the Ministry of Science and Technology (MOST) in Taiwan (Project: 106-2221-E-002-207-MY2 & 108-2221-E-002-104-MY3).